\documentclass[letterpaper]{article} 
\usepackage{aaai25}  
\usepackage{times}  
\usepackage{helvet}  
\usepackage{courier}  
\usepackage[hyphens]{url}  
\usepackage{graphicx} 
\usepackage{natbib}  
\usepackage{caption} 
\usepackage{amsmath}
\usepackage{algorithm}
\usepackage{algorithmic}
\usepackage{newfloat}
\usepackage{listings}
\usepackage{enumitem}
\usepackage{amsmath}
\usepackage{booktabs}
\usepackage{multirow}
\usepackage{tabularx}
\DeclareCaptionStyle{ruled}{labelfont=normalfont,labelsep=colon,strut=off} 
\floatstyle{ruled}
\newfloat{listing}{tb}{lst}{}
\floatname{listing}{Listing}

\title{Generative Adversarial Patches for Physical Attacks on Cross-Modal Pedestrian Re-Identification}
\author{
    Yue Su\textsuperscript{\rm 1}
    Hao Li\textsuperscript{\rm 1}
    Maoguo Gong\textsuperscript{\rm 1}
}
\affiliations{
    \textsuperscript{\rm 1} Key Laboratory of Cooperative Intelligent Systems, Ministry of Education, Xidian University\\

}

\usepackage{bibentry}

\begin{document}

\maketitle

\begin{abstract}
    Visible-infrared pedestrian Re-identification (VI-ReID) aims to match pedestrian images
    captured by infrared cameras and visible cameras. However, VI-ReID, like other traditional cross-modal image matching tasks, poses
    significant challenges due to its human-centered nature.
    This is evidenced by the shortcomings of existing methods, which struggle to extract common features across modalities,
    while losing valuable information when bridging the gap between them in the implicit feature space, potentially compromising security.
    To address this vulnerability, this paper introduces the first physical adversarial attack against VI-ReID models.
    Our method, termed Edge-Attack, specifically tests the model's ability to leverage deep-level implicit features by
    focusing on edge information - the most salient explicit feature differentiating individuals across modalities.
    Edge-Attack utilizes a novel two-step approach. First, a multi-level edge feature extractor is trained in a
    self-supervised manner to capture discriminative edge representations for each individual. Second, a generative model
    based on Vision Transformer Generative Adversarial Networks (ViTGAN) is employed to generate adversarial patches conditioned on
    the extracted edge features.By applying these patches to pedestrian clothing, we create realistic, physically-realizable adversarial samples.
    This black-box, self-supervised approach ensures the generalizability of our attack against various VI-ReID models.
    Extensive experiments on SYSU-MM01 and RegDB datasets, including real-world deployments, demonstrate the effectiveness of
    Edge-Attack in significantly degrading the performance of state-of-the-art VI-ReID methods.
\end{abstract}
\section{Introduction}
\begin{figure}[!t]
    \centering
    \includegraphics[width=\columnwidth]{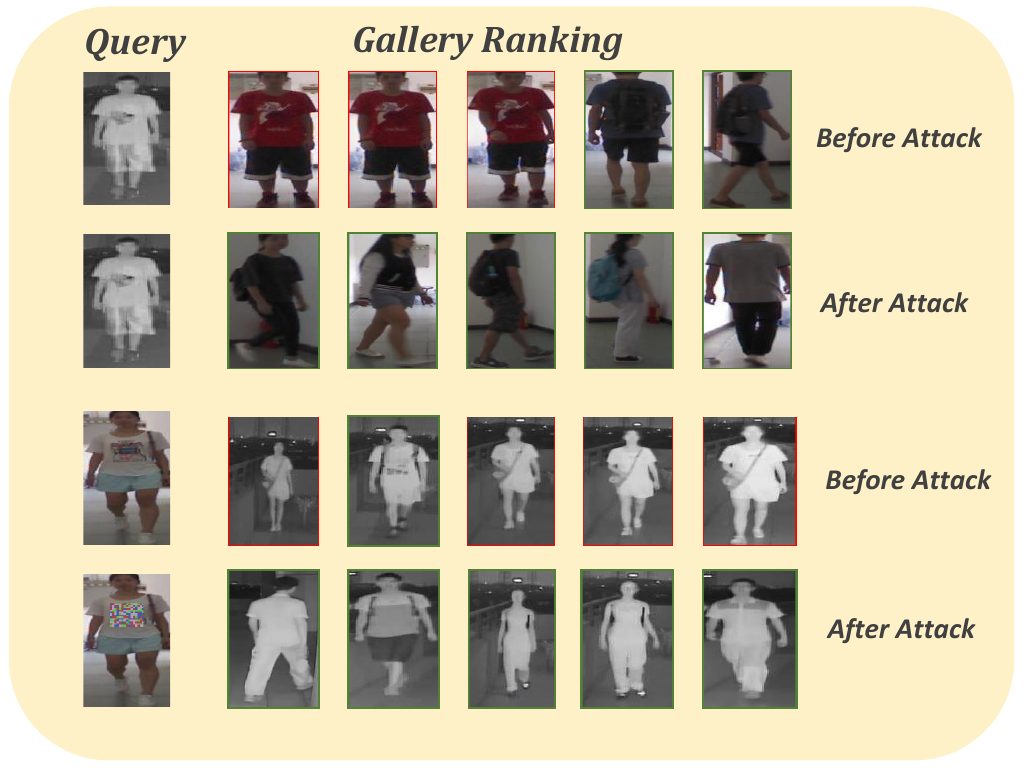}
    \caption{VI-ReID has two query methods: one is to match the queried pedestrian in the infrared scene from the visible scene, and the other is to match the queried pedestrian in the visible scene from the infrared scene. Generally speaking, the model will give the image sequence with the highest matching degree. Our method aims to make sure that there are no correctly matched targets in the model output sequence.}
    \label{fig1}
\end{figure}
Pedestrian re-identification is a task of finding target pedestrians in the huge amount of video frames acquired by a large network of
non-overlapping multi-view cameras, which can be regarded as a target retrieval problem. Traditional pedestrian re-identification is
based on RGB images, however, at night it is difficult for cameras to capture clear RGB images, and infrared photography is often
taken at this time. For example, in the pursuit of criminals in the task, many criminals are in the night to commit crimes, we often
can only get the criminals of the night action of the infrared image; at this time, if we want to re-find the whereabouts of criminals, we
must call other scenes of the camera shot by the RGB image and the infrared image shot at night, through the similarity scoring sorting
to lock the target person. Conversely, when the target person of our query is shot in an RGB scene, we can also supplement it with
an IR scene to query for its match in the IR scene. This cross-modal pedestrian re-identification is then called Visible-infrared Pedestrian
Re-identification(VI-ReID).

Inspired by deep learning, much of the existing ReID work is
based on DNNs, however, research has shown that DNNs are very
vulnerable and susceptible to attacks\cite{EOT}, which poses a considerable
security risk to our pedestrian re-identification work: a vulnerability in
the model often means a successful escape for the criminals. Among other pitfalls,
many attacks against deep ReID models have arisen, and these attacks have been shown to have the
following implications: (1)From the perspective of the monitoring
party, adversarial attacks help to find loopholes in the model. The
monitoring party can introduce adversarial samples to train the
model to make the model more robust to cope with emergencies.
(2)From the perspective of the monitored party, certain celebrities
need to protect their privacy to prevent harmful elements from
re-identifying them by intruding into the surveillance system and
obtain their whereabouts. Therefore, they need to process their images into adversarial samples to prevent re-identification.

As a result, many adversarial attacks against Re-ID have been
carried out in recent years. However, these works have many problems when faced with actual situations: (1)Except for one article
on physical adversarial attacks\cite{Advpattern}, most of the existing research on
ReID attacks are implemented in the digital field. Most of their attack methods are to perturb the pixels of the image itself. In fact,
it is difficult for the monitored object to have permission to obtain
information about the captured image, which means that our attack methods of perturbing the image itself are often ineffective
in a practical sense. However, physical adversarial attacks such as
designing adversarial patterns on pedestrians' clothes can ensure
that our adversarial samples can appear in any captured image. At
the same time, this attack mode can be implemented in black box,
so it also adapts to the fact that we can not know ID in advance as
model information requirements. On the other hand, people being
monitored are often unaware of when and where they are being
monitored by which camera mode. This also requires us to prepare adversarial samples in advance instead of processing images
in real time. This is only possible with physical adversarial attacks.
(2)Existing attacks all attack the RGB single-modality ReID task except for\cite{attackVI} in digital realm,
which means that when we use infrared images as a supplement to
match the target, most attacks fail; on the other hand, when infrared
images itself is used as the query target, attacks
against it is also a blue ocean. In this scenario, criminal suspects
who master cross-modal attacks against VI-ReID can easily
escape the pursuit of video surveillance.

All of the above urges us to develop a cross-modal adversarial attack method against VI-ReID,
and it is a physically achievable attack method. This requires us to take into account the
fragility of the existing VI-ReID model. The existing VI-ReID model mainly uses two methods
to make up for the information difference between modalities. The first is channel enhancement\cite{CAJ},
that is, trying to convert the RGB image into a grayscale image similar to the infrared image
through channel merging and then comparing it. This is proven to significantly waste the features
of the visible image. Another and more widely adopted method is to find the common feature space
of the two modalities\cite{DEEN,feng2023shape,AGW,DDAG,wang2019learning}. However, this implicit space often does not make full use of the common features and is easy to stay at the shallow
level of feature capture,
which is difficult to be discovered due to the implicit representation\cite{feng2023shape}.
This defect is difficult to quantify, which also forces
VI-ReID researchers to use external forces similar to graph learning to extract more deep-level modal-related
information that can be exploited in the implicit space as much as possible\cite{yang2024shallow,wu2023unsupervisedgraph,yang2023towards,cheng2023efficient}.
Based on the above situation, our method only exploits
the most significant and shallow common features of the two modalities: edge features for adversarial attacks. Specifically,
we designed an edge feature extractor and decoded the extracted edge features into adversarial patches through a
generative model and input them to the original image. Through self-supervision, it generates adversarial samples
with reverse edge features from the original image to attack VI-ReID models as shown in Figure\ref{fig1}.This method effectively tests whether the
common implicit feature extraction of the existing models still stays at the most shallow features
and its ability to mine deep-level features,
since the more the model stays at mining shallow edge features, the weaker its ability to resist attacks.
Our method can also provide samples for the future adversarial training of the model to guide the model to query more deep-level common features.

It is worth mentioning that in the entire field of physical adversarial attacks, we are the first to
use generative models to implement attacks. Compared with other widely used optimization-based methods,
this method can be deployed in advance under black-box conditions through self-supervision, so it can more
conveniently generate adversarial samples and can be more fully adapted to different models and scenarios.
This means that it better follows the principle of physical adversarial attacks being available in real-world scenarios.
The main contributions of our method are as follows:
\begin{itemize}
    \item We launched the first physical adversarial attack on the VI-ReID task. It is not a simple attack method, but aims to explore the ability of existing VI-ReID models to extract deep features and provide samples for correcting their feature extraction.
    \item We first introduced a generative model into physical adversarial attacks. Compared with previous optimization methods, this method allows physical adversarial attacks to generate samples in a black-box manner easy to deploy proactively to adapt to environmental changes. This greatly enhances the real-world versatility of physical adversarial attacks.
    \item We conducted attack experiments on various existing VI-ReID models on the SYSU-MM01 and RegDB datasets, and completed experiments in the physical domain. And these experiments have achieved remarkable results.
\end{itemize}

\section{Related Work}
\subsection{VI-ReID}
As mentioned above, VI-ReID refers to the task of searching for the same person between infrared and visible images.
\cite{firstVI} first proposed the VI-ReID problem and evaluated all the popular cross-domain models at that time.
Subsequently, many related works were carried out. Some of them focused on processing the channels of the image to achieve
alignment and matching of the modalities \cite{CAJ,wang2019rgb}, and some focused on finding the common feature space of the two
modalities \cite{DEEN,feng2023shape,AGW,DDAG,wang2019learning}.
However, relying solely on the above methods often cannot achieve sufficient matching of pedestrian IDs. In recent years,
VI-ReID has achieved better matching results by seeking unsupervised methods represented by graph learning,
etc. to more fully process common features \cite{yang2024shallow,wu2023unsupervisedgraph,yang2023towards,cheng2023efficient}.
\subsection{ Adversarial Attack on Re-ID}
Since the vulnerability of DNN was discovered, in the field of image recognition, many attacks against
RGB images\cite{hu2023physically,hu2021naturalistic}, attacks against
infrared images\cite{Wei_Wang_Jia_Zheng_Tang_Satoh_Wang_2023,hu2023adversarial}, and cross-modal attacks
targeting both modalities\cite{Wei_2023_ICCV,Dong_2019_CVPR} simultaneously have been discovered. Currently, adversarial
attacks against deep Re-ID mainly focus on the RGB field.\cite{Advpattern}implements physical adversarial attacks against deep
Re-ID for the first time, and designs two attack modes to evade
search and impersonate targets. \cite{MUAP} designed a universal
adversarial perturbation (MUAP) for the Re-ID model to attack the
model by disrupting the similarity ranking. \cite{rankattack} attack deep Re-ID systems by tricking them into learning incorrect
similarity rankings.
\cite{Metaattack} introduces a metalearning method in adversarial attacks to implement gradient interaction between multiple data sets to find general adversarial
perturbations. This perturbation is an overlapping part of multiple
data set fields, so the attack effect is widely efficient. \cite{MMattack}
is unique. The author uses adversarial attacks as a method of target protection. It sets the attack so that the probe's feature does
not match the target ID in the gallery and is different from other
IDs. By isolating the probe, it will not be queried by any probe
through the Re-ID model.\cite{Uturn} directly performs an
opposition-direction feature attack on the probe, and generates an
adversarial sample by completely pushing the probe's features in
the opposite direction, minimizing the probability of it being correctly matched in the gallery.
\cite{attackVI} first applied universal adversarial perturbations to VI-ReID, but as we described above, it lacks the general applicability of the physical world.
\section{Methodology}
\begin{figure*}[!t]
    \centering
    \includegraphics[height=0.52\textheight,width=\textwidth]{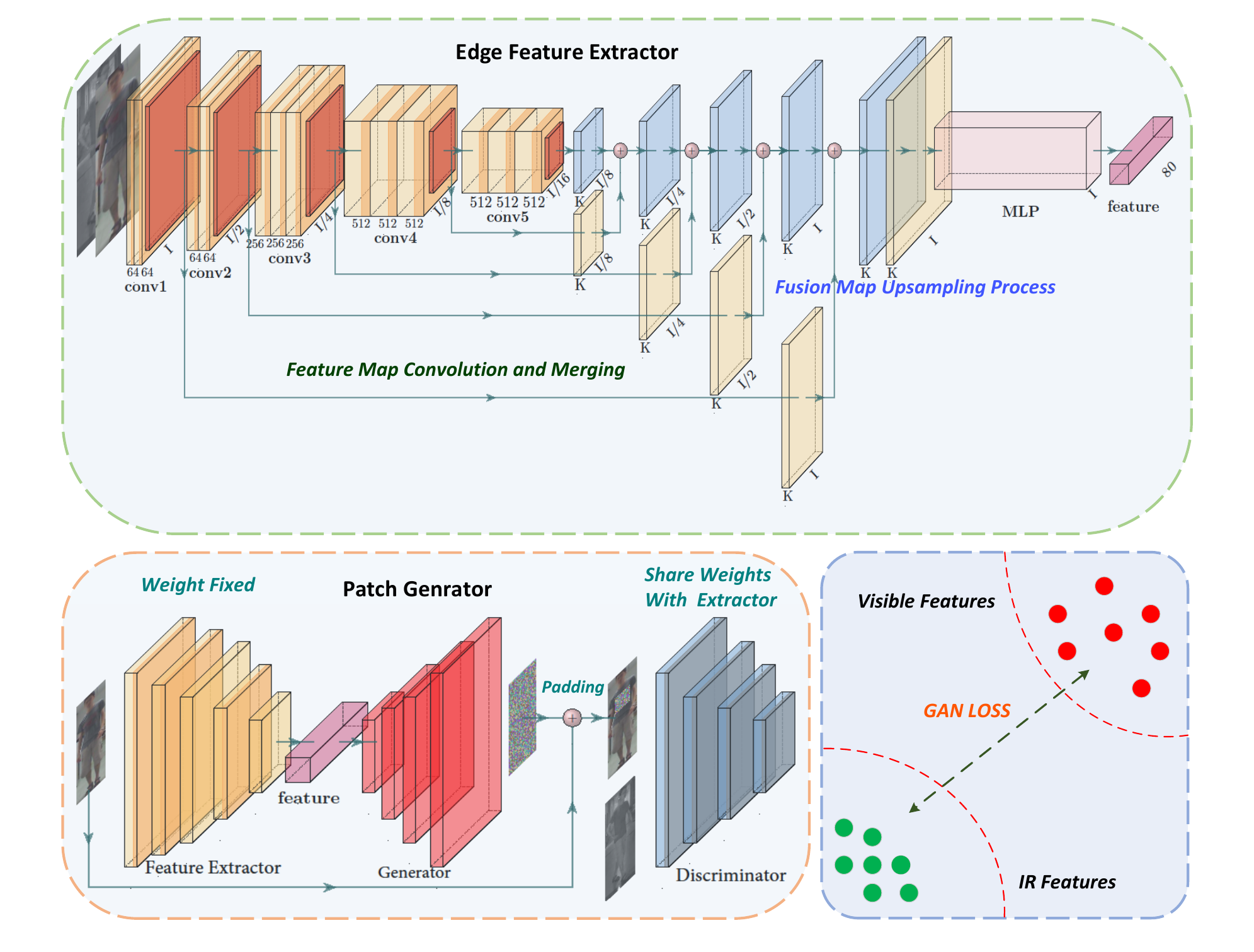}
    \caption{The figure shows an overview of the Edge-Attack method. The edge feature extractor obtains fine-grained edge features that can be used to distinguish pedestrian identities by fusing multi-level edge features. Our patch generator uses this feature to generate adversarial patches, and uses the feature extractor as a discriminator to generate reverse edge features, thereby creating our target adversarial sample.}
    \label{pipeline}
\end{figure*}
\subsection{Problem Definition}
We aim to attack a VI-ReID model, $f(\theta)$, by crafting an adversarial patch, $\eta$, that maximizes the
ranking of correctly matched images when applied to a visible image $V_p$ ($V'_p=V_p+\eta$).
Given a pedestrian $p \in P$, with corresponding infrared and visible image sequences $I_p$ and $V_p$, the attack is formulated as follows:
\begin{equation}
    max \sum_{p \in P} Rank_I(I_p) \ \ \ \ \ \ \ s.t. \ Rank_I=f(V'_{p},\theta)
\end{equation}
\begin{equation}
    max \sum_{p \in P} Rank_V(V'_p) \ \ \ \ \ \ \ s.t. \ Rank_V=f(I_{p},\theta)
\end{equation}
This effectively forces the model to misidentify the target pedestrian by pushing the correct matches to the bottom of
the ranking list. Our approach, illustrated in Figure \ref{pipeline}, leverages an edge feature extractor and a
patch generator to achieve this goal.
\subsection{Edge Feature Extraction}
Our model relies heavily on edge features for achieving fine-grained distinctions between pedestrians.
To ensure precise and robust edge feature extraction, we employ a multi-level fusion approach.

\textbf{Edge Detection:} We begin by extracting edge information from visible and infrared images using the classic HED Blocks \cite{HED},
a pre-trained network for edge detection. This process yields five edge maps, $L_1, L_2, ..., L_5$,
representing edge information at different levels of detail. These maps are visualized as orange squares in Figure \ref{pipeline}.

\textbf{Feature Refinement}: To further enhance edge feature representation, we process these five maps through a series of
convolutional blocks. Each map is processed independently, and then upsampled to restore its original size,
effectively preserving spatial context. This sequential upsampling operation is crucial for ensuring consistent feature
representation across different levels of detail. The final edge representation, $u_i$, is obtained by concatenating the upsampled
features as follows:
\begin{equation}
    u_i=U(Conv(L_{6-i})+u_{i-1}) \ \ \ s.t. \ 0<i<6
\end{equation}
where $u_i$ represents the feature map obtained at each stage of upsampling
(visualized as blue squares in Figure \ref{pipeline}). $U$ denotes the upsampling operation, and $Conv$ represents the convolution operation.

\textbf{Encoding:} Finally, the extracted features are passed through a convolution block and an MLP layer to generate the final edge features, $f_{edge}$. We refer to this process as edge feature encoding. Importantly, we use the same edge feature extractor for both infrared and visible images, ensuring consistency in our feature extraction process.

\textbf{Self-Supervised Training:} To effectively learn discriminative edge features, we train our feature extractor
in a self-supervised manner. The HED Blocks are fixed during training, and we focus solely on training the encoding part of the network.
This approach allows us to learn robust and discriminative edge features without requiring labeled data.
Specifically, we cluster the extracted features from all images based on their corresponding pedestrian ID.
This clustering process yields the final edge features, $f_{edge}^{V}$ and $f_{edge}^{I}$, for each pedestrian in the
visible and infrared modalities, respectively. Our objective function for training the feature extractor is formulated as follows:
\begin{equation}
    \begin{split}
        min \sum_{p \in P}\sum_{q \in P,q \ne p} \|f_{edge}^{V_p}-f_{edge}^{I_p}\|_2 - \\
        (\|f_{edge}^{V_p}-f_{edge}^{V_q}\|_2+\|f_{edge}^{I_p}-f_{edge}^{I_q}\|_2)
    \end{split}
\end{equation}
This objective function aims to maximize the difference between edge features extracted from the same
pedestrian in the visible and infrared modalities, while minimizing the difference between edge features
from different pedestrians within the same modality. This ensures that our edge features are highly discriminative
for identifying individuals across modalities.

\subsection{Adversarial Patch Generation}
To generate physically realizable adversarial patches, we leverage a generative model, a novel
approach in the realm of physical adversarial attacks. Unlike conventional optimization-based methods,
our generative model allows for black-box attack generation and facilitates proactive deployment in real-world scenarios.

\textbf{Generative Model Selection:} We adopt ViT-GAN \cite{vitgan} as our generative framework.
This architecture utilizes Vision Transformers (ViT) for image generation, mapping image labels in
pixel space to generate images. Our choice of ViT-GAN instead of widely used DDPM\cite{DDPM} or VAE\cite{VAE} is motivated
by the following advantages:

\textit{Effective Edge Feature Decoding:} ViT-GAN excels at decoding edge features, especially in the early stages of generation. This capability is crucial because, even after feature fusion, explicit features alone may not fully distinguish all pedestrian identities. The ViT generator effectively embeds these features into a low-dimensional space, facilitating the design of highly effective adversarial patches.

\textit{Robust Training:} The incorporation of a Generative Adversarial Network (GAN) architecture enhances training robustness, enabling better control over patch stability and reverse feature generation. Although this comes at the expense of feature space learning, the impact is minimal in our case due to the single nature of edge features.

\textit{Natural Patch Generation:} ViT-GAN's unique patch mechanism generates samples with a natural appearance. While not as uniform as diffusion models, this mechanism aligns perfectly with our goal of generating non-semantic patches, ensuring they blend seamlessly with clothing and appear unobtrusive in real-world settings.

\textbf{Patch Generation Process:} Specifically, we retain the generator of ViTGAN. For each pedestrian $p$,
we first input its visible images $V_p$ into the trained edge feature extractor.
The extracted edge features are then used as input to the ViT generator, which decodes them into an adversarial patch $\eta$:
\begin{equation}
    \begin{split}
        \eta & = \text{Decoder}_{\text{ViTGAN}}(\text{MLP}_2(\text{MLP}_1([\mathbf{z}, \mathbf{y}])) + \\
        &\text{PositionalEncoding}(\text{MLP}_1([\mathbf{z}, \mathbf{y}]))) \\
        &\quad \text{where} \quad \mathbf{z} \sim \mathcal{N}(0, \mathbf{I})
    \end{split}
\end{equation}
Where $\mathbf{z}$ represents the latent vector of ViTGAN, and $\mathbf{y}$ denotes the embedded feature vector,
which in our case corresponds to the edge feature vector $f_{edge}^{V_p}$ output by the extractor.
We utilize the edge feature extractor as the discriminator within our
ViT-GAN framework. After applying the generated patch $\eta$ to the pedestrian image, resulting in $V'p$,
we input both $V'p$ and the corresponding infrared images $I_p$ into the discriminator.
This process generates edge features for both modalities, which are then clustered based on pedestrian ID
to obtain $f{edge}^{V'}$ and $f{edge}^{I}$. Our loss function is defined as follows:
\begin{equation}
    \mathcal{L} =  \frac{1}{\|P\|}\sum_{p \in P} \|f_{edge}^{V'_p}-f_{edge}^{I_p}\|_2
\end{equation}
This loss function encourages the generator to create patches that maximize the difference between edge
features of the same person in different modalities.
Notably, our training process is self-supervised and black-box, as it does not require any feedback from the
VI-ReID model being attacked. This ensures the generalizability of our approach to various VI-ReID models.
\subsection{Attack}
\begin{algorithm}[htbp]
    \caption{Edge-Attack}
    \label{alg:attack}
    \begin{algorithmic}
        \REQUIRE Trained feature extractor $E$, ViT generator $G$
        \REQUIRE Visible image $V_p$ VI-ReID model $f(\theta)$, Infrared Gallery Set $I$
        \STATE $f_{edge} = E(V_p)$ \COMMENT{Extract edge features}
        \STATE $\eta = G(f_{edge})$ \COMMENT{Generate adversarial patch}
        \STATE $V_p' = V_p + \eta$ \COMMENT{Apply patch to image}
        \STATE $Rank_I = f(V_p', \theta)$ \COMMENT{Get ranking from the model}
        \RETURN $Rank_I$ \COMMENT{Matching results}
    \end{algorithmic}
\end{algorithm}
The deployment of the Edge-Attack, as outlined in Algorithm \ref{alg:attack},
involves a straightforward procedure. Once the edge feature extractor ($E$) and the ViT-based patch generator ($G$)
are adequately trained, an attack can be executed efficiently. Initially, the visible image ($V_p$) of
the target pedestrian is fed into the trained edge feature extractor to obtain the corresponding edge features ($f_{edge}$).
Subsequently, these extracted features are used as input to the ViT generator, which produces the adversarial patch ($\eta$).
The patch is then applied to the original visible image, creating the adversarial sample ($V_p'$). Finally,
the adversarial sample is presented to the targeted VI-ReID model ($f(\theta)$), and the resulting ranking ($Rank_I$) is observed.
This process effectively demonstrates the attack against the VI-ReID model when using the visible modality as the query.
The procedure remains consistent when using the infrared modality as the query.
This streamlined process highlights the practicality of Edge-Attack.
The separation of the training phase and the attack deployment phase allows for efficient generation of adversarial samples.
By leveraging the pre-trained components, Edge-Attack can readily craft adversarial samples for any given visible or infrared image,
enabling a comprehensive evaluation of the targeted VI-ReID model's vulnerability.
\section{Experiments}
\begin{table*}[!t]
    \centering
    \renewcommand\arraystretch{1.6}
    \setlength\tabcolsep{1.6pt}
    \footnotesize
    \begin{tabularx}{\textwidth}{ccccccccccccccccc}
        \toprule
        \multirow{2}{*}{Method} & \multicolumn{4}{c}{SYSU-MM01 (ALL)} & \multicolumn{4}{c}{SYSU-MM01 (INDOOR)} & \multicolumn{4}{c}{RegDB (VIS to IR)} & \multicolumn{4}{c}{RegDB (IR to VIS)}                                                                                                 \\
        \cmidrule(lr){2-5}\cmidrule(lr){6-9}\cmidrule(lr){10-13}\cmidrule(lr){14-17}
                                & r=1                                 & r=10                                   & r=20                                  & mAP                                   & r=1   & r=10  & r=20  & mAP   & r=1   & r=10  & r=20  & mAP   & r=1   & r=10  & r=20  & mAP   \\
        \midrule
        $AGW$\cite{AGW}         & 47.50                               & 84.39                                  & 92.14                                 & 47.65                                 & 54.17 & 91.14 & 95.98 & 62.97 & 70.05 & 86.21 & 91.55 & 66.37 & 70.49 & 87.21 & 91.84 & 65.90 \\
        $AGW(After Attack)$     & 1.13                                & 11.20                                  & 21.82                                 & 2.95                                  & 1.10  & 11.44 & 22.35 & 3.01  & 1.31  & 7.52  & 11.80 & 2.11  & 0.87  & 6.07  & 9.66  & 1.77  \\
        \midrule
        $CAJ$ \cite{CAJ}        & 69.88                               & 95.71                                  & 98.46                                 & 66.89                                 & 76.26 & 97.88 & 99.49 & 80.37 & 85.03 & 95.49 & 97.54 & 79.14 & 84.75 & 95.33 & 97.51 & 77.82 \\
        $CAJ(After Attack)$     & 1.10                                & 11.36                                  & 22.25                                 & 3.00                                  & 1.10  & 11.28 & 22.25 & 3.01  & 1.02  & 6.94  & 10.68 & 1.74  & 0.87  & 6.07  & 9.61  & 1.76  \\
        \midrule
        $DDAG$ \cite{DDAG}      & 54.75                               & 90.39                                  & 95.81                                 & 53.02                                 & 61.02 & 94.06 & 98.41 & 67.98 & 69.34 & 86.19 & 91.49 & 63.46 & 68.06 & 85.15 & 90.31 & 61.80 \\
        $DDAG(After Attack)$    & 0.89                                & 11.20                                  & 21.77                                 & 2.95                                  & 1.77  & 19.66 & 38.13 & 6.93  & 3.83  & 11.84 & 17.52 & 4.99  & 4.61  & 13.25 & 18.16 & 4.28  \\
        \midrule
        $DEEN$ \cite{DEEN}      & 74.70                               & 97.60                                  & 99.20                                 & 71.80                                 & 80.30 & 99.00 & 99.80 & 83.30 & 91.10 & 97.80 & 98.90 & 85.10 & 89.50 & 96.80 & 98.40 & 83.40 \\
        $DEEN(After Attack)$    & 1.45                                & 11.78                                  & 22.22                                 & 3.07                                  & 1.86  & 19.84 & 37.86 & 7.07  & 1.59  & 18.61 & 38.77 & 6.63  & 1.49  & 19.52 & 37.68 & 6.71  \\
        \bottomrule
    \end{tabularx}
    \caption{Performance of four VI-ReID models before and after adversarial attacks using Edge-Attack. The results are shown in terms of Rank-r (\%) and mAP (\%). We evaluated on two datasets: SYSU-MM01 (with indoor and full scene search) and RegDB (with VIS to IR and IR to VIS search modes). }
    \label{atr}
\end{table*}
\begin{table*}[!t]
    \centering
    \renewcommand\arraystretch{1.6}
    \setlength\tabcolsep{3.6pt}
    \footnotesize
    \begin{tabular}{ccccccccccccccccc}
        \hline
        \multirow{2}{*}{Removed Levels} & \multicolumn{4}{c}{SYSU-MM01 (ALL)} & \multicolumn{4}{c}{SYSU-MM01 (INDOOR)} & \multicolumn{4}{c}{RegDB (VIS to IR)} & \multicolumn{4}{c}{RegDB (IR to VIS)}                                                                                                 \\
        \cline{2-17}
                                        & r=1                                 & r=10                                   & r=20                                  & mAP                                   & r=1   & r=10  & r=20  & mAP   & r=1   & r=10  & r=20  & mAP   & r=1   & r=10  & r=20  & mAP   \\
        \hline
        $None$                          & 1.13                                & 11.20                                  & 21.82                                 & 2.95                                  & 1.10  & 11.44 & 22.35 & 3.01  & 1.31  & 7.52  & 11.80 & 2.11  & 0.87  & 6.07  & 9.66  & 1.77  \\
        $L_5$                           & 5.44                                & 11.36                                  & 16.07                                 & 6.62                                  & 6.11  & 13.59 & 19.37 & 8.24  & 4.66  & 13.06 & 17.23 & 4.70  & 5.49  & 13.30 & 16.99 & 6.75  \\
        $L_4, L_5$                      & 11.36                               & 19.61                                  & 23.50                                 & 11.54                                 & 19.22 & 32.38 & 40.49 & 19.69 & 13.01 & 29.27 & 40.39 & 15.04 & 12.86 & 25.87 & 34.85 & 14.42 \\
        $L_3, L_4, L_5$                 & 21.33                               & 66.50                                  & 82.43                                 & 23.24                                 & 17.25 & 55.04 & 72.13 & 18.27 & 27.14 & 51.31 & 62.28 & 28.63 & 21.26 & 45.49 & 57.57 & 22.92 \\
        $L_2, L_3, L_4, L_5$            & 38.79                               & 83.30                                  & 92.95                                 & 40.04                                 & 34.24 & 75.52 & 86.77 & 33.39 & 34.13 & 60.63 & 71.31 & 34.48 & 38.11 & 66.70 & 76.12 & 37.55 \\
        $All$                           & 44.57                               & 83.04                                  & 92.19                                 & 43.29                                 & 53.77 & 87.67 & 95.24 & 53.53 & 54.33 & 88.51 & 94.74 & 53.19 & 63.98 & 82.77 & 88.66 & 55.55 \\
        \hline
    \end{tabular}
    \caption{Ablation study of removing edge feature maps. Rank-r (\%) and mAP (\%) are reported for different datasets and removal levels. The \textbf{All} represents removing all five layers. }
    \label{Ablation}
\end{table*}
In this section, we present our experimental process, which includes attacks on various
existing VI-ReID models on two datasets, as well as real-world physical attack experiments.

\textbf{Datasets}
Our method is evaluated using two prominent cross-modality ReID datasets: SYSU-MM01\cite{firstVI} and RegDB\cite{RegDB}.
SYSU-MM01 is a comprehensive dataset featuring 395 identities captured by 6 cameras (4 RGB and 2 near-infrared) on the
SYSU campus. It contains 22,258 visible images and 11,909 near-infrared images. For testing, the dataset includes 95
identities, with queries drawn from 3803 images taken by two IR cameras. We conduct ten evaluation runs following\cite{firstVI}.
RegDB is a more compact dataset with 412 identities, each represented by ten visible and ten thermal images.
We assess performance in two retrieval settings: from visible tothermal and from thermal to visible.

\textbf{Evaluation Metrics}
We use common metrics for person re-identification to evaluate the effectiveness of the VI-ReID model
and our attack. Specifically, we use the rank-r method, which refers to the probability of the target
person being in the confidence ranking of the matching images given by the model when we restrict ourselves
to taking only the first r images.In addition, we use mAP(mean Average Precision) as the evaluation indicator of average precision\cite{mAP}.

\textbf{The Model Under Attack}
We selected four methods for attack. Among them, AGW\cite{AGW} and DDAG\cite{DDAG} are widely used baselines
in cross-modal person re-identification tasks, and they are both feature-based methods. CAJ\cite{CAJ}
is a sota method among the current channel-based methods. DEEN\cite{DEEN} is a feature-based sota method.
CAJ compensates for the difference in visible and infrared modes by randomly using color channels to generate
grayscale images, so precisely arranged perturbation attacks are also unstable.DEEN searches for the widest cross-modal
common feature space through a clever feature space embedding design, which is what we are most interested in.
If the feature coverage of the implicit space is wider, then the attack effect of using only the reverse edge
feature will be worse, and vice versa. Therefore, we are eager to explore whether the common feature space searched by
the current VI-ReID methods is wide.
\subsection{Attack Performance}
The experimental results are shown in Table \ref{atr}.We have achieved good results in different scenarios and modal search methods.
It can be found that we have reduced the accuracy of most methods to around 1\% in rank 1 search, and reduced the mAP of all methods to below 7\%.
It is worth noting that our method works best on AGW and channel-based CAJ, which have not undergone proprietary feature space
processing. This also proves that the common features they searched for are the coarsest. The average mAP of all datasets of DDAG
and DEEN remained at around 5\% after the attack, which proves that although they have mined certain
cross-modal identity features, a large part of the features still remain on the shallow edge features,
without more effective extraction of fine-grained features, so they do not show high robustness to reverse
edge feature attacks. The results also reflect that these two methods may be able to search for specified people
in the lower rankings, so we infer that the non-edge features they extracted may also lack high independence for different identities.

More generally speaking, the matching between the two modalities is still highly dependent on shallow
features under existing deep learning models, which is very disadvantageous for practical scenarios.
It is difficult to interfere with this feature learning preference simply by changing the architecture of the implicit space.
Our sample generation method can be used by VI-ReID researchers to train the model more robustly in the future.
\subsection{Ablation Study}
In this section, we mainly conduct ablation experiments on the edge feature extractor. We remove the
5-level feature maps of the network one by one and explore the contribution of edge features at different
levels to the attack effect. Our attack is performed on the AGW baseline. The experimental results are shown in table\ref{Ablation}.

From the results, we can see that mAP changes significantly with the removal of each layer of edge feature maps,
especially after all edge feature maps are removed (which means that the patch is generated solely by the global features
extracted from the original image through the convolutional block), mAP increases by nearly 20\% compared to retaining
$L_1$, and at this time mAP is already quite close to the non-attack state (see Table\ref{atr}). This means that attacks that
generally rely on global features to generate patches are ineffective.

We can also see that when $L_3$ is removed, the growth of rank-10 and rank-20 is the most significant, which
means that the attack effect is weakened after removing $L_3$ in real-world applications is the most obvious.
This layer is often the layer that eliminates the noise of environmental edge features and retains all edge features of
the human body to the greatest extent. This confirms that our reverse attack on edge features is effective. In addition,
this phenomenon also shows that the VI-ReID model is relatively robust to changes in scene information when extracting features.
It is also quite sufficient for extracting edge features of the human body instead of just focusing on local parts. However,
it lacks effective methods for deep features extract.
\subsection{Real-world Experiment}
\begin{figure}[!t]
    \centering
    \includegraphics[width=\columnwidth]{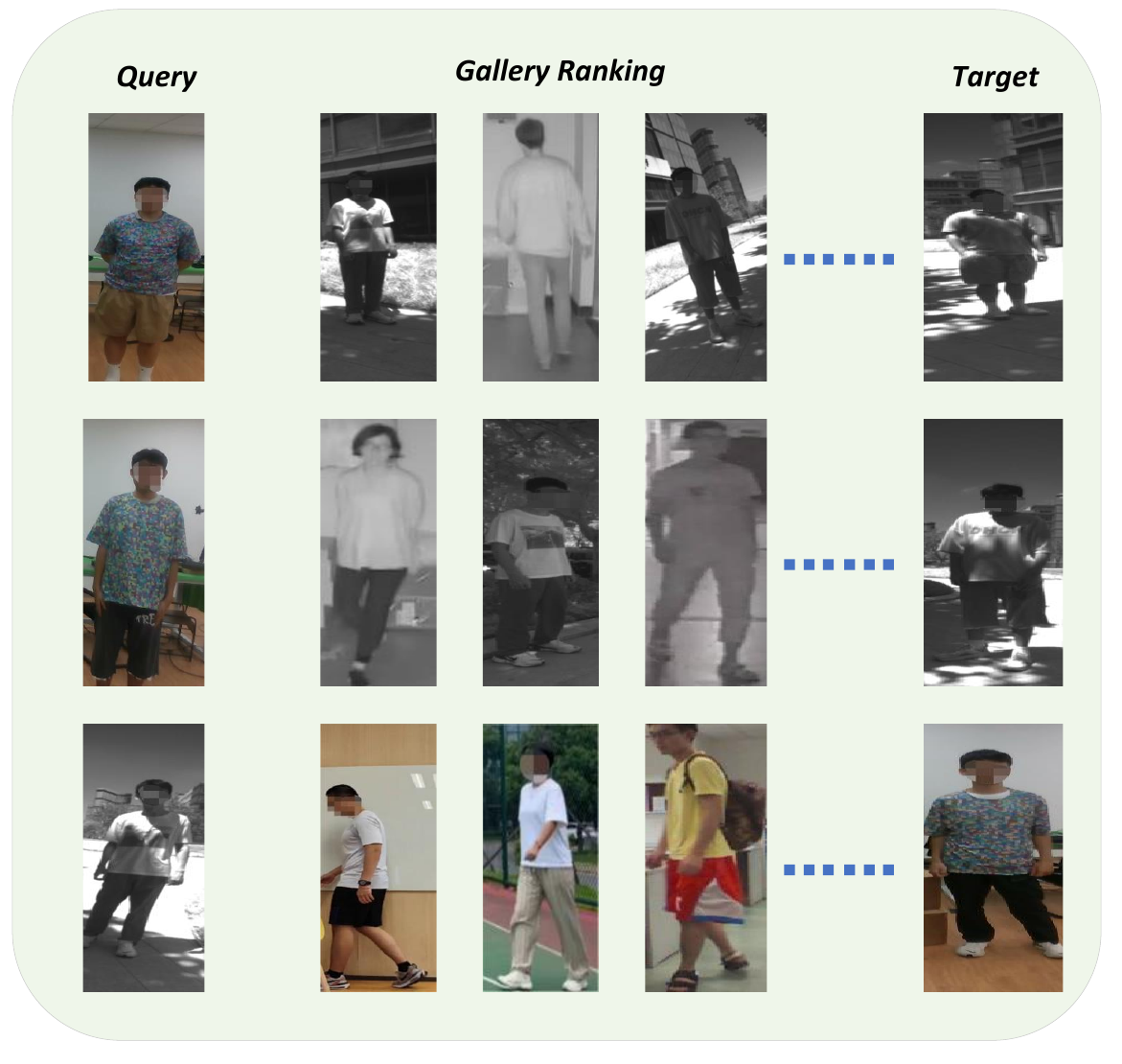}
    \caption{Our physical experiment scene setting. We design the adversarial IDs into clothes by using patches generated by Edge-Attack. IDs wearing these clothes will experience two modes of search on the AGW baseline: VIS TO IR (first two rows) and IR TO VIS (last row). The model will give a matching sequence, and our experiments have proven that it can make the ranking of the correctly matched IDs lag far behind, causing a powerful attack on the VI-ReID model.}
    \label{phy}
\end{figure}
\begin{table}[!t]
    \centering
    \renewcommand\arraystretch{1.0}
    \setlength\tabcolsep{10pt}
    \footnotesize
    \begin{tabular}{ccccc}
        \toprule
                    & \multicolumn{2}{c}{IR to VIS} & \multicolumn{2}{c}{VIS to IR}             \\
        \cmidrule(l){2-3}\cmidrule(l){4-5}
        $ID$        & r=1                           & r=5                           & r=1 & r=5 \\
        \midrule
        $1$         & 80                            & 100                           & 70  & 100 \\
        $1(Attack)$ & 0                             & 10                            & 0   & 15  \\
        $2$         & 85                            & 100                           & 75  & 100 \\
        $2(Attack)$ & 0                             & 5                             & 0   & 15  \\
        $3$         & 85                            & 100                           & 70  & 100 \\
        $3(Attack)$ & 0                             & 10                            & 0   & 5   \\
        \bottomrule
    \end{tabular}
    \caption{Comparison of three IDs before and after physical adversarial attacks on the AGW baseline in two search modes. We report Rank-r (\%) as the evaluation metric.}
    \label{atr1}
\end{table}
Based on the digital domain experiments, we conducted physical adversarial attacks in the real world.
By using patches generated by generative model in the digital domain to make camouflage-style clothing,
we truly achieved an adversarial style that is realistic and wearable by people.

\textbf{Experiment Setup}  Our experiment set up 3 IDs for attack. Due to the limited number of people,
for each query, we selected 6 IDs taken by ourselves and 14 IDs added from the SYSU-MM01 dataset as pedestrian gallery.
And set 5 RGB images and 5 infrared images for each ID for query. We selected AGW baseline as the attacked model.
In order to ensure the unbiasedness of the model, we pre-trained it on the RegDB dataset. In the scene setting, we set
indoor and outdoor scene images for each ID, and took different shooting angles and shooting distances in each image.
During the attack process, we first tested the query accuracy of the selected ID in two query modes without wearing our
designed adversarial clothes, and then tested the query accuracy of the selected ID wearing our adversarial clothes.
We used 20 tests. In each test, we replaced the IDs from ourselves and the dataset at a ratio of 50\% to enrich the
gallery setting and more fully evaluate the attack effect. In terms of the selection of indicators, due to the small total
number of our images, we selected rank-1 and rank-5 as evaluation indicators.

\textbf{Experiment Results}  The experimental results are shown in the table\ref{atr1}. Our physical adversarial attack results
have achieved great success in both modal searches. In the IR to VIS search mode, our attack reduces the model's rank-1
and rank-5 indicators by an average of 83.3\% and 91.7\%. In the VIS to IR search mode, our attack reduces the model's rank-1
and rank-5 indicators by an average of 71.7\% and 88.3\%. This fully demonstrates the high usability and versatility of our
method in the physical world.
Furthermore, the success of Edge-Attack in a real-world scenario highlights its practical implications for security systems 
that rely on VI-ReID. The ability to generate realistic adversarial patches that can be physically implemented on clothing 
showcases the potential for real-world attacks against such systems, which are more challenging to simulate and defend against 
compared to purely digital attacks. This underscores the importance of considering physical adversarial attacks when designing 
robust VI-ReID systems.
\section{Conclusion}
In this article, we propose a physical adversarial attack method for the cross-modal person re-identification task.
It generates adversarial samples through reverse edge features to test the ability of existing cross-modal person
re-identification models to extract feature spaces. A large number of experiments have proven the significant effect of
this method, which also illustrates that the existing VI-ReID model lacks the ability to extract deep-level modal common
features. In addition, this paper applies the generative model to physical adversarial attacks for the first time, making
it easier for physical adversarial attacks to generate adversarial samples first without relying on model feedback,
which increases the versatility and portability of the attack. The limitation of this article is that an adversarial
training method that uses adversarial samples generated by Edge-Attack to optimize the feature search direction of existing
VI-ReID has not yet been developed. This will also be our future research direction.
\bibliography{aaai25}
\end{document}